\documentclass[runningheads,table]{llncs}
\usepackage[dvipsnames]{xcolor}
\usepackage{eccvabbrv}
\usepackage{wrapfig}
\usepackage{tikz}
\usepackage{bbding}
\usepackage{pifont}
\usepackage{comment}

\usepackage{eccv}
\usepackage{graphicx}
\usepackage{booktabs}
\usepackage{multirow}
\usepackage{pifont}
\usepackage{textcomp}
\usepackage[accsupp]{axessibility} 
\usepackage[pagebackref,breaklinks,colorlinks,citecolor=eccvblue]{hyperref}
\usepackage{hyperref}

\begin{document}

\title{MobilePortrait: Real-Time One-Shot
 Neural Head Avatars on Mobile Devices} 

\titlerunning{MobilePortrait}
\authorrunning{Jiang  et al.}

\author{
Jianwen Jiang\thanks{Equal Contribution.} \and
Gaojie Lin$^{\star}$ \and
Zhengkun Rong \and
Chao Liang \and \\
Yongming Zhu \and
Jiaqi Yang \and
Tianyun Zhong
}

\institute{ByteDance Inc. \\
\email{jianwen.alan@gmail.com, lingaojie@bytedance.com}} 

\maketitle

\begin{figure}
    \centering
    \includegraphics[width=1.0\linewidth]{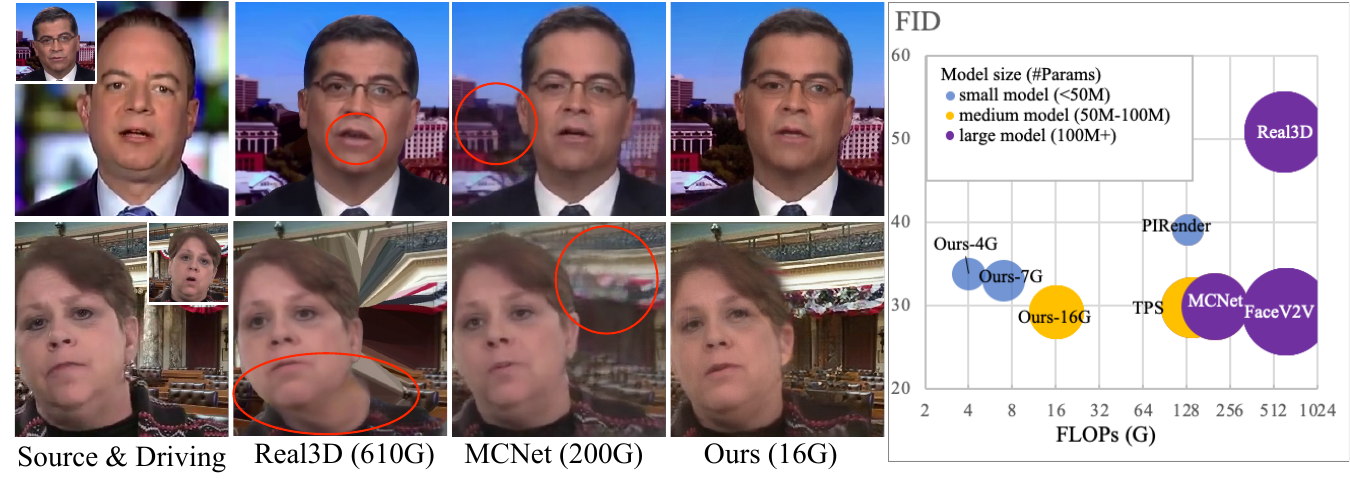}

    \caption{The provided examples (on the left) demonstrate that our methods can achieve results comparable to or even better than those of current high-computation state-of-the-art methods, but with less than one-tenth of the computational cost. On the right, a bubble chart compares various methods, with the size of each bubble representing the model's parameter size. This further confirms that our method can produce high-quality results while offering a significant advantage in computational efficiency.
    }
    \label{fig:firstpage}
\end{figure}
\begin{abstract}
Existing neural head avatars methods have achieved significant progress in the image quality and motion range of portrait animation. However, these methods neglect the computational overhead, and to the best of our knowledge, none is designed to run on mobile devices. This paper presents MobilePortrait, a lightweight one-shot neural head avatars method that reduces learning complexity by integrating external knowledge into both the motion modeling and image synthesis, enabling real-time inference on mobile devices. Specifically, we introduce a mixed representation of explicit and implicit keypoints for precise motion modeling and precomputed visual features for enhanced foreground and background synthesis. With these two key designs and using simple U-Nets as backbones, our method matches state-of-the-art performance with less than one-tenth the computational demand. It has been validated to reach speeds of over 50 FPS on mobile devices and support both video and audio-driven inputs. Video samples are in \href{https://mobileportrait.github.io/}{project page}.
\end{abstract}

\section{Introduction}
\label{sec:intro}
One-Shot Neural head avatar (NHA) is a technology that animates a single facial image according to a specified driving signal, which could be video or audio, to synthesize a portrait video. In recent years, significant improvements~\cite{drobyshev2022megaportraits, zhang2023metaportrait,wang2021facev2v,zhao2022tps,hong2023mcn,ye2024real3d,ren2021pirenderer,yin2022styleheat,wang2022lia} have been made in the quality of the generated image and the range of motion. However, existing NHA approaches concentrate on achieving realism and robustness in image synthesis with models that are increasingly complex, typically surpassing 100 GFLOPs~\cite{ye2024real3d,wang2021facev2v,zhao2022tps,hong2023mcn}, leading to the under-exploration of lightweight NHA. With the swift progress in large language models (LLMs) and the widespread use of smartphones, avatars on mobile devices are poised to become a crucial interface for AI interaction. This prospect has driven us to develop an efficient one-shot neural head avatar model optimized for performance on mobile platforms.
\par
Initially, we attempted to convert existing SOTA models into ones that could be deployed on mobile devices. However, we found that these models incorporated many complex modules in their structural design, such as memory modules, dynamic convolutions~\cite{hong2023mcn}, attentions~\cite{hong2023mcn,mallya2022implicit}, multiscale feature warping~\cite{hong2023mcn,wang2021facev2v,zhao2022tps} or image-to-plane~\cite{ye2024real3d} methods. Reducing computational complexity is a challenging task, and the complexity of the models increased the difficulty and development workload of deploying them on mobile devices. Therefore, we start to reflect on the rationale underlying these methods and aim to construct a lightweight NHA model through the most essential and straightforward design.

\par

In fact, Real3D~\cite{ye2024real3d} and MCNet~\cite{hong2023mcn} represent two distinct categories of motion modeling: explicit facial movement modeling and implicit global motion modeling. Explicit modeling~\cite{zhang2023metaportrait,ye2024real3d,ren2021pirenderer} methods often involve predefined facial keypoints or 3D face representation to capture motion driven by facial movements. This results in undefined motion in regions beyond the face, necessitating a powerful motion network to extrapolate and fill in the motion for these areas based solely on facial movements. Implicit modeling methods~\cite{wang2022lia,zhao2022tps,siarohin2021mraa,siarohin2019fomm,hong2023mcn} use an encoder to extract global and image-level motion from inputs without facial priors, representing motion with neural keypoints or latents, which requires a powerful motion network to define facial and background movements.

From the Figure~\ref{fig:firstpage}, we can observe that the explicit modeling method, Real3D, produces poor results in areas not defined by the 3DMM, such as inside the mouth and around the neck. Meanwhile, the implicit modeling method, MCNet, produces notable blurriness at the boundary between the person and the background, possibly due to the lack of an explicit facial region prior.
This observation inspired us to develop a more holistic and efficient approach to motion modeling that combines face-specific knowledge with global motion representations, complementing each other.
\par
The results shown in Figure~\ref{fig:firstpage} not only reveal issues with motion capturing but also indicate inadequate appearance synthesis capability of the model.
Many recent advancements~\cite{shi2023mvdream,jiang2019mlvcnn,su2015multi,morf,chen2017multi} in 3D-related fields have been driven by the integration of multiview designs into network architectures. The rationale is intuitive: a network exposed to more appearances can learn more effectively. This concept has led us to explore a parallel avenue: if facial knowledge can strengthen our motion network, might the incorporation of appearance knowledge similarly strengthen our synthesis network?
The integration of appearance knowledge may redefine the synthesis network's task, transforming it from generating all content with a powerful network to efficiently completing content with the provided appearance knowledge, akin to shifting from a closed-book to an open-book exam. Importantly, this can be achieved with virtually no increase in computational load during runtime because appearance knowledge can be prepared in advance.
\par
Building on aforementioned observation and considerations, we meticulously design MobilePortrait, our lightweight one-shot neural head avatar method. First, we utilize lightweight U-Nets with conventional convolutional layers as the backbones for motion and synthesis networks, significantly reducing computational requirements compared to existing methods and is easily implementable on mobile devices. Second, to compensate for potential losses in motion accuracy due to reduced computations, we combine implicit global motion modeling with explicit facial motion modeling, introducing mixed keypoints to capture motion. We also design facial knowledge losses to ensure the incorporation of facial knowledge. Lastly, in the image synthesis phase, we incorporated appearance knowledge, utilizing pseudo multi-view features and pseudo backgrounds to enhance synthesis of foreground and background respectively. With these proposed designs, MobilePortrait achieves performance on par with or exceeding state-of-the-art methods with far less computational demand, as shown in Figure~\ref{fig:firstpage}. Our contributions are succinctly outlined as follows:

\begin{itemize} 
\item We introduce MobilePortrait, which, to the best of our knowledge, is the first one-shot mobile neural head avatar method capable of real-time performance.
\item We streamline the task by leveraging external facial and appearance knowledge, merging explicit and implicit keypoints for comprehensive motion capture, and including features like pseudo multiview and background for improved synthesis. This approach allows MobilePortrait to efficiently create neural head avatars with lightweight U-Net~\cite{unet} backbones.
\item Extensive testing across various datasets confirms MobilePortrait's effectiveness. It achieves state-of-the-art performance while requiring significantly fewer FLOPs and parameters. Moreover, we have also verified that MobilePortrait can render at speeds of up to 100+ FPS on mobile devices and support both video and audio driving inputs.
\end{itemize}

\section{Related Works}
\label{sec:related}

Neural head avatar generation can be categorized into video-driven and cross-modal driven approaches. video-driven neural head avatars methods mainly consists of two important parts: motion modeling and image synthesis. The former captures the motion between the source and driving images, while the latter generates the animated pixels.
For motion modeling, some methods \cite{siarohin2019fomm,siarohin2021mraa,zhao2022tps,hong2023mcn, wang2021facev2v, wang2022lia} propose frameworks for decoupling appearance and motion representation in an unsupervised manner.
For instance, \cite{siarohin2019fomm,siarohin2021mraa,zhao2022tps,hong2023mcn} involve learning to detect 2D implicit keypoints from images and further predict explicit warping flow. FaceV2V~\cite{wang2021facev2v} expands the network architecture dimension and learns 3D implicit keypoints for motion modeling.
LIA~\cite{wang2022lia} constructs a latent motion space and represents motion as a linear displacement of the latent code.
In contrast, other works \cite{zhang2023metaportrait,ren2021pirenderer,ye2024real3d} rely on explicit motion representation, such as pre-defined facial landmarks and blendshapes.
For example, MetaPortrait~\cite{zhang2023metaportrait} uses facial landmarks as input to predict a warp flow, while PIRenderer~\cite{ren2021pirenderer} and Real3D~\cite{ye2024real3d} employ the 3DMM model~\cite{song20093dmm} to facilitate decoupling of the control of facial rotation, translation and expression. Although significant progress has been made, implicit and explicit modeling still remain largely independent of each other.
For image synthesis, some methods \cite{siarohin2019fomm, siarohin2021mraa, zhao2022tps} predict motion flow to directly warp the source image, utilizing more original pixel information, while other approaches \cite{wang2021facev2v, zhang2023metaportrait, hong2023mcn, drobyshev2022megaportraits} opt to warp features, offering greater flexibility for subsequent generative networks. StyleHeat~\cite{yin2022styleheat} explores using pretrained StyleGAN~\cite{karras2019stylegan} as a generator, achieving neural head avatars through latent edits on the powerful generator.

Current cross-modal methods, mainly audio-driven, aim to generate motion signals from audio for natural talking head videos with accurate lip-sync and expressive facial animations.
They typically produce driving signals as output and use separately trained video-driven models as video renderers.
Sadtalker~\cite{zhang2023sadtalker} uses a PoseVAE and ExpNet to generate head pose and expression as motion descriptors generated from audio and adopt FaceV2V~\cite{wang2021facev2v} as renderer. Vivid-Talk~\cite{sun2023vividtalk} designs an Audio-To-Mesh module to predict 3DMM expression coefficients and lip-related vertex offsets based on an input audio signal and a reference image, while utilizing another mesh-to-video model as a renderer. Some recent works~\cite{he2023gaia,ma2023dreamtalk,zhang2023dream,he2023gaia,ma2023dreamtalk,zhang2023dream} explore the use of diffusion models as audio-to-motion modules, employing VAEs~\cite{kingma2013auto} or existing video-driven models as renderers to enhance the accuracy and expressiveness of motion signals. EMO~\cite{tian2024emo} designs a end-to-end diffusion model and can generate highly realistic portrait videos based on audio input. Although the inference process is end-to-end, the multi-stage training procedure for the network can, to some extent, correspond to the audio-to-motion and render modules.Image quality in audio-driven methods largely hinges on the rendering model or the module managing image quality. Given the driving signal's low transmission cost, it's suitable for server-side deployment. Thus, an efficient renderer is key for audio-driven neural head avatars on edge devices.

In general, existing works typically pursue complex network architectures and high computational complexity to achieve high-quality animation but rarely consider scenarios with limited computational resources, leaving on-device neural head avatars generation largely unexplored.

\begin{figure}[t]
    \centering
    \includegraphics[width=1.0\linewidth]{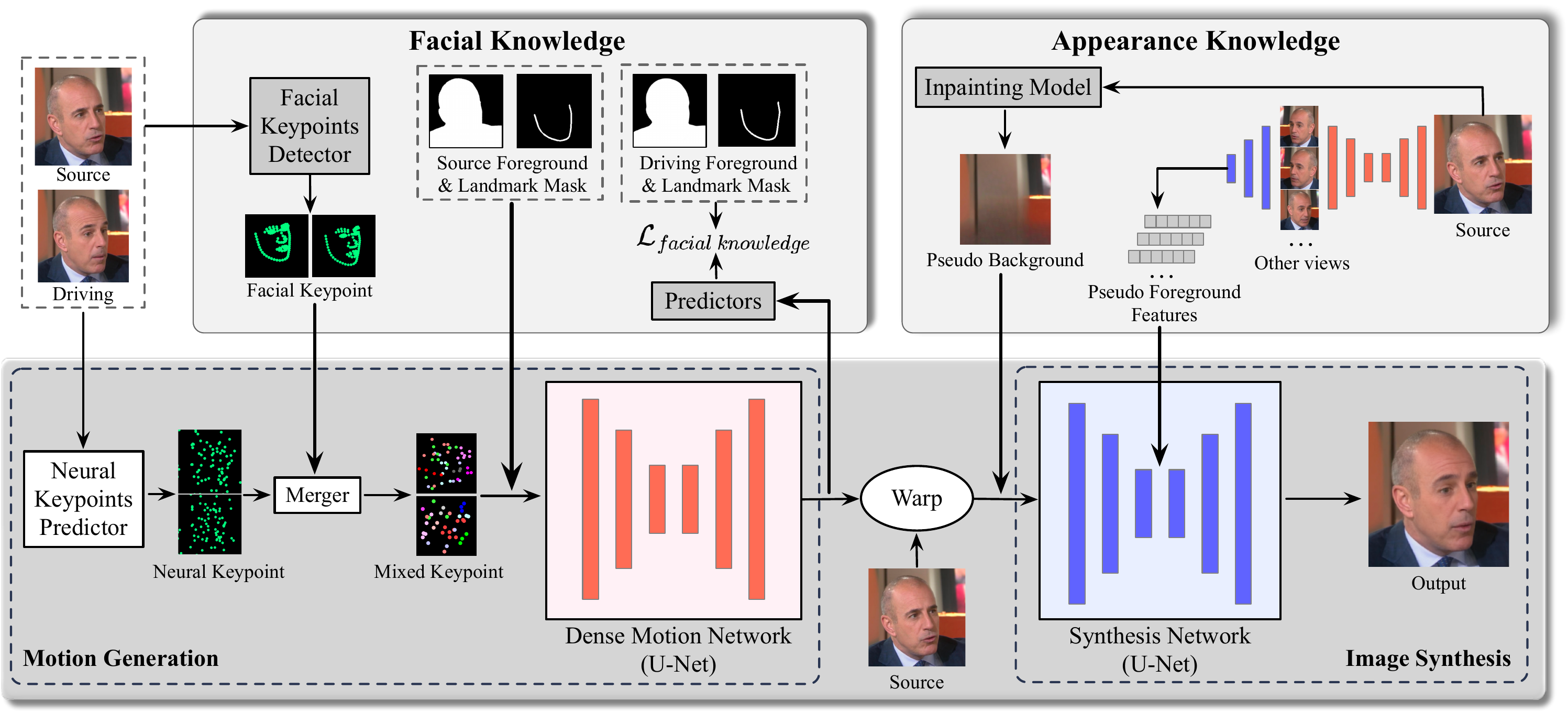}

    \caption{
    \textbf{The video-driven pipeline of MobilePortrait.} MobilePortrait processes source and driving image to generate mixed keypoints that are merged from detected neural and facial keypoints. These mixed keypoints, along with precomputed source masks, are used to create optical flow for image warping via a dense motion network. The synthesis network generates the final image by combining the warped image with precomputed pseudo background and multiview foreground features. Since facial and appearance knowledge is precomputed just once, the two simple U-Net backbones account for nearly all of the computational load during inference. In audio-driven mode, an audio-to-keypoints module supplies the driving keypoints.}
    \label{fig:framework}
\end{figure}
\section{Method}
\label{sec:method}
This section first provides an overview of MobilePortrait's architecture, shown in Figure~\ref{fig:framework}, comprising two primary modules: motion generation and image synthesis. Then in Section~\ref{sec:motion} we describe the hybrid motion modeling designed within the motion generation module, which utilizes both explicit and implicit facial keypoints. Next, we introduce techniques that enhance image synthesis through precomputed appearance knowledge in Section~\ref{sec:synthsis}. Subsequently, in Section~\ref{sec:mobile}, we present the audio-to-motion module, which allows MobilePortrait to be driven by audio input. Finally, we outline the loss functions employed during training in Section~\ref{sec:training}.
\subsection{Overview of MobilePortrait}
\label{sec:overview}
As depicted in Figure~\ref{fig:framework}, with video-driven animation as an example, MobilePortrait processes the source image $\mathbf{S}$ and each driving frame $\mathbf{D}$ from the driving video, generating target images frame by frame. Specifically, within the motion generation module, Keypoint Detectors initially produce a set of keypoints, which are our proposed mixed keypoints in the MobilePortrait, for both $\mathbf{S}$ and $\mathbf{D}$, \ie $\{x_{s,i}, y_{s,i}\}_{i=1}^{N_{mk}}$ and $\{x_{d,i}, y_{d,i}\}_{i=1}^{N_{mk}}$. The subsequent warping and generation process is similar to the previous works~\cite{zhao2022tps,hong2023mcn,siarohin2019fomm,siarohin2021mraa}. Based on these keypoints, we follow TPS~\cite{zhao2022tps} to generate the initial transformations. Keypoints represented as heatmaps are input into the dense motion network and combined with initial transformations to generate the motion field, $\mathbf{M}$, delineating the pixel displacement from $\mathbf{S}$ to $\mathbf{D}$, or in other words, the optical flow. Based on the source image and the optical flow, a warp operation is performed to obtain the initial warped image, which is then multiplied by another output from the dense motion, the occlusion maps, to produce the final warped image $\mathbf{S}_{w}$. Subsequently, the Image Synthesis module leverages $\mathbf{S}_{w}$ and auxiliary appearance knowledge features derived from $\mathbf{S}$ to create the final target image through a synthesis network. For efficient computation and mobile deployment friendliness, we retained simple U-Nets without the additions from prior work~\cite{hong2023mcn, zhao2022tps, wang2021facev2v}, such as multi-scale feature warping, dynamic convolution, and attention modules, as backbones for both the dense motion network and the synthesis network.

\begin{figure}[t]
    \centering
    \includegraphics[width=1.0\linewidth]{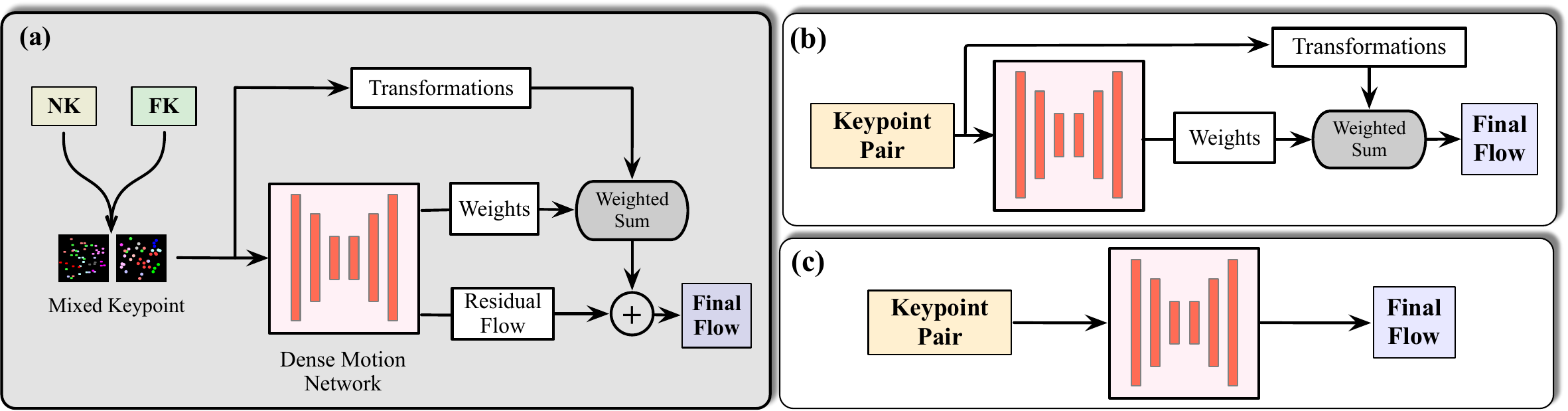}

    \caption{
    \textbf{The motion generation process of MobilePortrait.} (a) represents the optical flow generation method adopted by our MobilePortrait, where NK and FK represent the neural and facial keypoints, respectively. (b) is the method used in literatures~\cite{siarohin2019fomm,zhao2022tps,hong2023mcn,siarohin2021mraa}; (c) is similar to literature~\cite{zhang2023metaportrait}, which directly obtains optical flow through CNN. For brevity, we omitted the heatmap generation and occlusion process.
    }
    \label{fig:demo_motion}
\end{figure}
\subsection{Motion Generation with Facial Knowledge}
\label{sec:motion}
\textbf{Mixed Keypoint Representation.} In the Motion Generation module, prior works such as FOMM~\cite{siarohin2019fomm}, TPS~\cite{zhao2022tps}, and MCNet~\cite{hong2023mcn} employ similar network design structures. They utilize a neural keypoint predictor, denoted as NK detector, to separately predict a pair of keypoints for $\mathbf{S}$ and $\mathbf{D}$, and based on these keypoints, construct an initial collection of transformations. Dense motion network (DMN) then predicts local weights for this collection of transformations and occlusion maps for warped image. The optical flow field is obtained through a weighted summation of these elements. This process is similar to part (b) described in Figure~\ref{fig:demo_motion}.
\par
Neural keypoints enable the network to learn global motion information from the driving video, as well as facial movements. However, as the computational load of the dense motion network decreases, the network struggles to distinguish between the motion of the face and the background, leading to severe artifacting, akin to a "liquefaction" effect, or may even result in an inability to drive the synthesized video, as shown in the visualization results in Figure~\ref{fig:gflops_vis}. 
To address this, we introduce a pretrained face keypoint detector to extract facial landmarks from $\mathbf{S}$ and $\mathbf{D}$ respectively. A mixed keypoint predictor, the merger shown in Figure~\ref{fig:framework}, then merges the neural keypoints and the face keypoints to create mixed keypoints. As shown in the left part of Figure~\ref{fig:demo_motion}, once the mixed keypoints are calculated, we proceed to calculate the optical flow based on these keypoints, replacing the neural keypoints used in previous methods~\cite{hong2023mcn,zhao2022tps,siarohin2019fomm,siarohin2021mraa}. Our experiments indicate that integrating implicit and explicit keypoints effectively reduces global liquefaction artifacts and enhances motion precision in the generated videos and also performs better than other methods of incorporating facial information. Additionally, inspired by MetaPortrait~\cite{zhang2023metaportrait} and the ResNet~\cite{he2016deep} architecture, we add two extra output channels to the last layer of our dense motion network. This modification enables the network to produce a residual optical flow, enhancing the expressiveness of the generated optical flow.

\par
\textbf{Face-Aware Motion Generation.} In addition to incorporating facial priors into the keypoints representation, we enrich the input to the dense motion network with a foreground mask and facial landmarks mask from the source image. These only need to be computed once, preserving real-time inference capabilities. As shown in Figure~\ref{fig:framework}, we further design a facial knowledge loss. Specifically, we add two predictors for these masks to the last feature layer of the DMN, which are trained with L1 losses to predict the foreground and landmarks mask for the driving image. These predictors, existing only during training, help the model to better understand portrait integrity, facilitating improved face-aware motion generation.
\par
With these enhancements, our motion generation module leverages external facial knowledge to perform motion capture at both the face level and the video level, with virtually no increase in computational cost. This enables the model to generate a plausible optical flow $\mathbf{M}$ even when the computational load of dense motion network is reduced.

\subsection{Image Synthesis with Appearance Knowledge}
\label{sec:synthsis}
Image synthesis based on the warped source image capitalizes on the original pixel information, but as the warping itself doesn't create new pixel data, reliance on the warped source may lead to diminished synthesis quality when there are changes in pose angles, as shown in Figure~\ref{fig:firstpage}. 
To compensate for the decrease in synthesis quality due to reduced complexity, we utilize the warped source image as input to U-Net based synthesis network and introduce precomputed visual features from source image to decrease the burden on  Image Synthesis module.
\par
\textbf{Enhanced Foreground Synthesis.} We sample $T$ frames uniformly from the driving video and, with the source image, generate $T$ newly warped images using our motion generation module. As depicted in the top-right part of Figure~\ref{fig:framework}, 
To ensure efficient feature extraction and fusion, we opt for the final downblocks of the U-Net, corresponding to the lowest spatial resolution. The early layers of the U-Net, up to the last downblock, are utilized for feature extraction from the newly warped image to obtain multiview features. An additional convolution layer merges multiview features with those of the current frame within the corresponding downblock. Apart from this, there are no further differences or additional computational burdens imposed on the synthesis network. These pseudo multiview image features offer appearance information from different poses to aid in enhancing the quality of synthesis and can be precomputed, thus not hindering inference efficiency. 
\par
\textbf{Enhanced Background Synthesis.} We employ an offline inpainting model to fill in the source image after foreground removal, creating a complete background picture as shown in the top-right part of Figure~\ref{fig:framework}. This inpainted background, along with a mask of the foreground, serves as extra inputs to the synthesis network. To ensure that the Image Synthesis module can effectively utilize this background information, we perform inpainting on the driving image during training, which has proven crucial in our experiments.
\par
With these improvements, our image synthesis can rely on a simple yet efficient U-Net backbone while maintaining high-quality synthesis results during inference, with negligible additional computational cost.
\subsection{Audio-Driven Functionality}
\label{sec:mobile}
In this section, we first introduce a baseline solution that enables MobilePortrait to support audio-driven functionality. To enable MobilePortrait to process audio-driven signals, we need to extract neural keypoints and facial keypoints from the audio input. Inspired by the audio-to-motion designed in SadTalker~\cite{zhang2023sadtalker} and VividTalk~\cite{sun2023vividtalk}, we train an audio-to-motion model that includes two modules: audio-to-mesh and mesh-to-neural keypoints. The former uses LSTM to convert audio signals into 3D Morphable Model (3DMM) coefficients to acquire facial meshes, whereas the latter employs a ResNet18~\cite{he2016deep} to predict neural keypoints from images sketched with sampled mesh vertices and edges. Facial keypoints are directly extracted from the mesh. With this setup, we capture the necessary motion signals, including neural and facial keypoints, for driving MobilePortrait with audio input. 
Thanks to the trained mesh-to-neural keypoints module, MobilePortrait can also be driven by 3DMM. This not only facilitates expression editing via 3DMM but also enhances results in cross-identity scenarios when driven by 3DMM. It is important to note that we provide merely a baseline solution here, enabling audio-driven capability for MobilePortrait. MobilePortrait can accommodate more sophisticated designs~\cite{sun2023vividtalk, ye2024real3d, zhang2023dream, ma2023dreamtalk} to achieve improved results.
\par

\subsection{Training Losses}
\label{sec:training}
Following previous works\cite{siarohin2019fomm,zhao2022tps,hong2023mcn}, 
We employ perceptual loss $\mathcal{L}_{percep}$ and L1 loss $\mathcal{L}_{L1}$ to optimize feature and pixel distances, keypoint distance loss $\mathcal{L}_{kp}$ for facial keypoints accuracy, and equivariance loss~\cite{zhao2022tps} $\mathcal{L}_{eq}$ for neural keypoint stability. Additionally, we add two proposed facial knowledge loss terms (shown in the top-left part of Figure~\ref{fig:framework}), implemented in L1 loss, to make dense motion network to be aware of the landmark mask $\mathcal{L}_{landmark}$ and foreground mask $\mathcal{L}_{mask}$.
The final loss can be written as follows:
\begin{equation}
    \mathcal{L} = \mathcal{L}_{percep}
    +\mathcal{L}_{L1}
    +\mathcal{L}_{kp}
    +\mathcal{L}_{eq} 
    +\mathcal{L}_{landmark}
    +\mathcal{L}_{mask}
\end{equation}
\section{Experiments}
\label{sec:exp}
\textbf{Experimental Setup}.
To rigorously assess method effectiveness, we trained and tested our approach using various datasets. For training, we leveraged the VFHQ~\cite{xie2022vfhq}, VoxCeleb2~\cite{vox}, and CelebvHQ~\cite{zhu2022celebv} datasets, which together comprise 16,827 high-resolution portrait clips from VFHQ, 35,666 clips at 512×512 from CelebVHQ, and 150,480 clips at 256×256 from VoxCeleb2, collectively representing more than 21,000 distinct identities. To evaluate generalization, We construct a test set drawn from multiple datasets, including Talking Head 1K~\cite{wang2021facev2v}, CCv2~\cite{ccv2}, and HDTF~\cite{hdtf}, from which we randomly sampled 38, 137, and 100 videos according to the dataset proportions, respectively. Videos were processed at 25 FPS, square-cropped based on face detection, and resized to 512px. 

\textbf{Implementation Details}. For facial keypoints FK, we adopt the 106 landmark protocol, and for neural keypoints NK, we select 50 points. The mixed keypoint predictor is realized by concatenating keypoints and processing them through a MLP. By fusing FK and NK, we obtain 50 mixed keypoints. A foreground segmenter~\cite{cheng2021mask2former} was employed for mask extraction, and LaMa~\cite{lama} was used for background inpainting. The training processes of MobilePortrait are performed on 8 NVIDIA A100 GPUs, with a learning rate of 0.002 for 60 epochs.

\textbf{Metrics} 
To comprehensively evaluate the efficacy of our method, we employed multiple metrics and assessed both same-id reenactment and cross-id reenactment. To evaluate the quality of the generated images, we used common image quality metrics~\cite{hong2023mcn,ye2024real3d} including reference-based indices like FID, SSIM, and PSNR, as well as identity preservation indicator CSIM. To measure the accuracy and stability of the synthesized motion, we evaluated average keypoint distance(AKD), head pose distance(HPD) and expression errors(AED). Additionally, referencing recent text-to-video evaluation metrics~\cite{huang2024vbench}, we add a background consistency index (BCI).

\textbf{Compared methods.} 
To validate the superiority and effectiveness of our method, we conducted comparative tests with recent top-performing methods, including latent-driven TPS, MCNet and FaceV2V, as well as approaches that use landmarks and 3DMM like Real3D, PIRender. For fair comparisons, we trained all methods on the same datasets as previously described, with the exception of Real3D. Due to its complex training requirements, we use the official release model trained on CelebVHQ. To account for this, we also included the performance of our method under the same conditions in Table~\ref{tab:dataset}. Notably, these methods were not initially designed for computational efficiency, which often results in them having higher FLOPs and parameter counts, making them challenging to deploy on mobile devices.
\begin{table}[t]\scriptsize
	\centering
	\caption{Comparisons with SOTA methods in video-driven same/cross-identity reenactment. \textbf{Bold} means best scores and \underline{underline} means top3 scores.}
    \setlength\tabcolsep{0.01mm}{
    \begin{tabular}{l|ccccccc|cccc|c}
        \toprule
        \multirow{2}{*}{Method} &  
        \multicolumn{7}{c|}{Same-Identity Reenactment  } & \multicolumn{4}{c}{Cross-Identity} &Cost
        \\
        & FID$\downarrow\rm\ $ &  PSNR$\uparrow\rm\ $ & SSIM$\uparrow\rm\ $ & AKD$\downarrow\rm\ $& HPD$\downarrow\rm\ $ & AED$\downarrow\rm\ $ & BCI$\uparrow\rm\ $ 
        & CSIM$\uparrow\rm\ $ & HPD$\downarrow\rm\ $ & AED$\downarrow\rm\ $ & BCI$\uparrow\rm\ $ &FLOPs(G)  \\ 
        \midrule
        
        PIRender~\cite{ren2021pirenderer} & 39.1 & 22.7 & 77.8  & 2.14 & 0.99 & 0.09 &96.9& 45.7 & 4.50 & 0.15&96.7 &131\\
        $\text{FaceV2V}$~\cite{wang2021facev2v} & 29.3 & 22.5 & 85.3 & 1.96 & 2.52 & 0.06  &97.2& 46.0 & 5.45 & 0.15&97.2&629 \\
        TPS~\cite{zhao2022tps} & 29.8 & 27.3 & 87.7 &  1.43 & 0.71 & 0.06 &97.9& 38.9 & 4.61 & 0.15 &97.6&140\\
        MCNet~\cite{hong2023mcn} & 27.2 & 28.5 & 88.7 & 1.33 & 0.81 & 0.05 &97.8& 27.6 & 6.69 & 0.16&97.5 &200\\
        Real3D~\cite{ye2024real3d} & 50.8 & 23.1 & 80.6 & 1.63 & 0.82 & 0.08&97.6 & 47.8 & 3.74 & 0.17 &97.5&610\\
        \rowcolor{gray!20} 
        Ours  &\underline{29.2} & \underline{26.1} & \underline{85.9} & \textbf{1.30} & \textbf{0.40} & \textbf{0.05}&\textbf{98.2} & 39.2 & \textbf{2.74} & \textbf{0.13} & \textbf{97.9}&\textbf{16}\\
        
        \bottomrule
    \end{tabular}}
    \label{tab:main_exp}
\end{table}

\begin{table}[t]\scriptsize
    \caption{\textbf{Experimental results of models with differernt FLOPs.}
    }
    \begin{minipage}{0.35\textwidth}%
        \setlength\tabcolsep{0.2mm}{
        \begin{tabular}{l|ccc}
            \toprule
            FLOPs & Device &\#Param. &\rm Latency \\
            \midrule
            16G & iPhone14 Pro &67.7M &15.8ms 
            \\
            7G & iPhone14 Pro &40.8M &6.4ms \\
            4G & iPhone14 Pro &25.5M &5.9ms \\
            \midrule
            16G & iPhone12&67.7M &25.5ms  \\
            7G & iPhone12&40.8M &10.9ms \\
            
            4G & iPhone12&25.5M &8.9ms \\

    	    \bottomrule
        \end{tabular}
        }
    \end{minipage}
    \hspace{1.0em}
    \begin{minipage}{0.6\textwidth}%
    \includegraphics[width=1\linewidth]{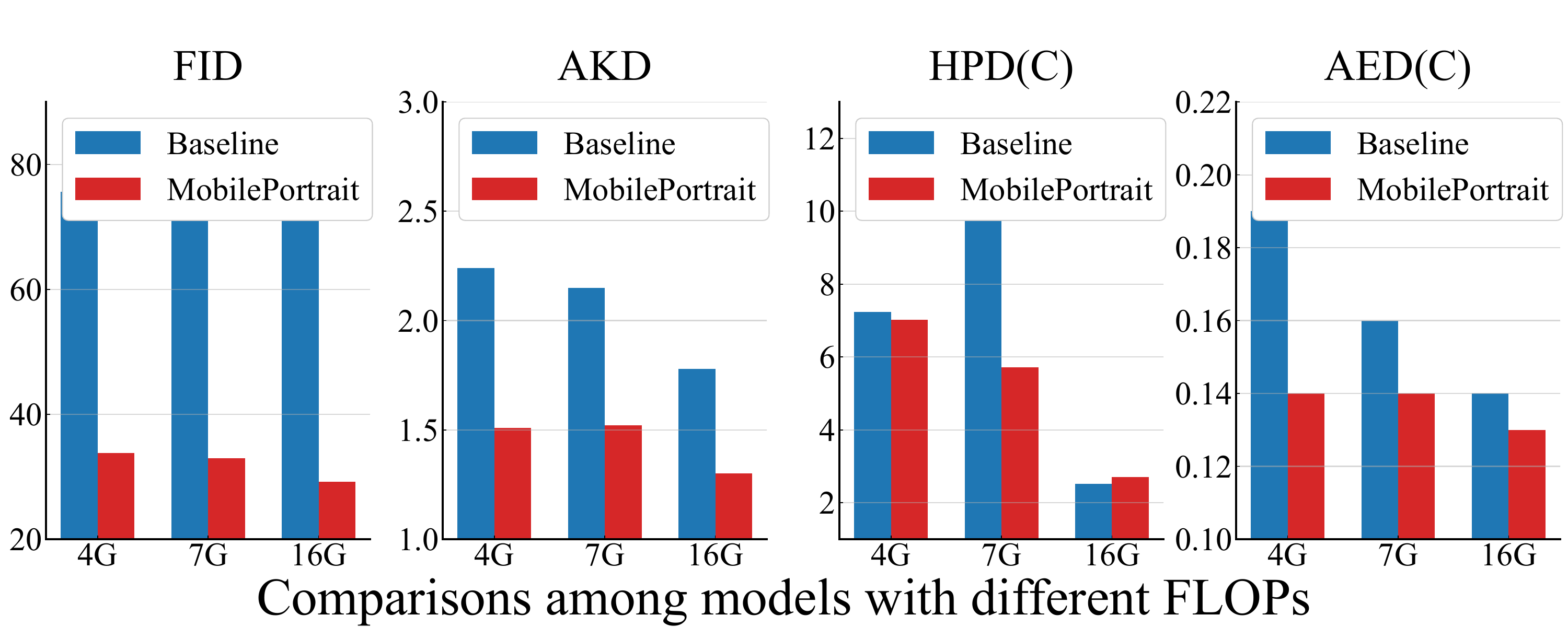}
    \end{minipage}
    
    \label{tab:gflops_exp}
\end{table}

\begin{figure}[t]
    \centering
    \includegraphics[width=1.0\linewidth]{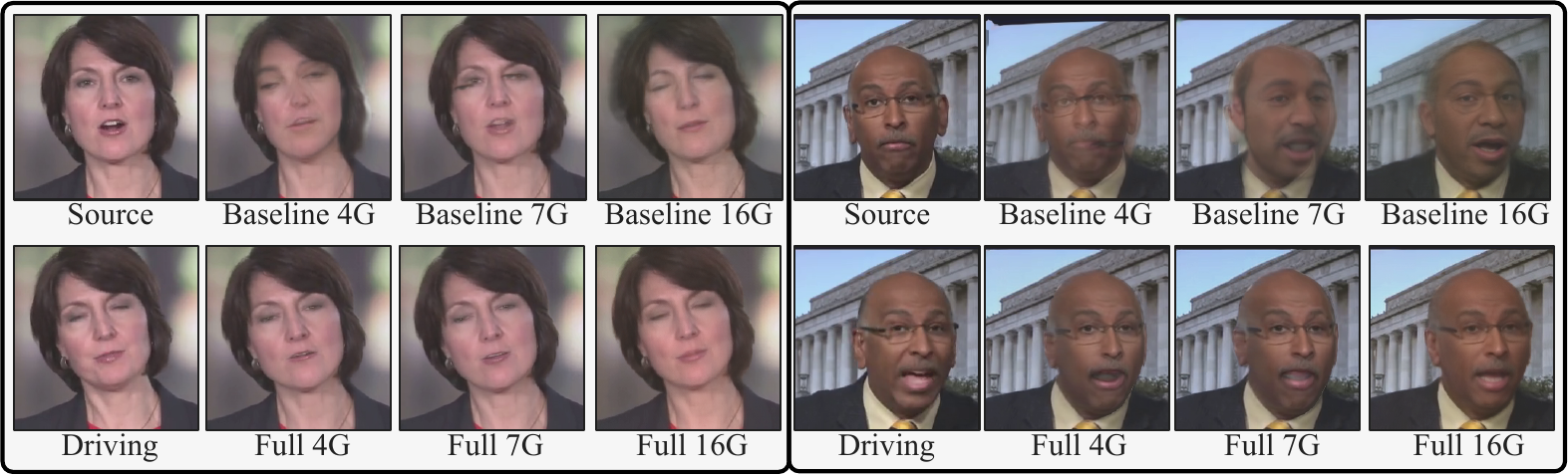}

    \caption{
    \textbf{Visualizations Comparisons among models with different FLOPs.} 
    }
    \label{fig:gflops_vis}
\end{figure}
\begin{table}[t]\scriptsize
	\renewcommand{\arraystretch}{1.1}
	\caption{\textbf{Ablation studies of motion generation.}
	}
	\begin{minipage}{0.4\textwidth}%
	\centering
    \setlength\tabcolsep{0.6mm}
    \begin{tabular}{c|cccc}
        \toprule
        Method & ${\rm FID}$& ${\rm AKD}$& ${\rm AED(C)}$& ${\rm HPD(C)}$ \\
	    \midrule
        \rowcolor{gray!20}
         Mixed Keypoint. & 29.2 & 1.30 & 3.0 & 0.13 \\ 
         NK-Only & 48.3 & 2.62 & 3.9& 0.17  \\ 
         FK-Only & 33.2 & 1.61 & 10.5& 0.13  \\ 
         No Proposed Loss & 29.1 & 1.45 & 4.19 & 0.13  \\ 
         
	    \bottomrule
     
    \end{tabular}
    \end{minipage}
    \hspace{4.0em}
    \begin{minipage}{0.3\textwidth}%
    \setlength\tabcolsep{0.6mm}
        \begin{tabular}{c|cccc}
        \toprule
        Method & ${\rm FID}$& ${\rm AKD}$& ${\rm AED(C)}$& ${\rm HPD(C)}$ \\
	    \midrule
        \rowcolor{gray!20}
         Ours & 29.2 & 1.30 & 3.0 & 0.13 \\ 
         No Residual O.F. & 28.5 & 1.45 & 6.2 & 0.13  \\ 
         Conv. Motion & 43.9 & 1.49 & 3.7& 0.15  \\ 
         Spase Motion & 44.7 & 3.14 & 9.5& 0.14  \\ 
         
	    \bottomrule
    \end{tabular}
    \end{minipage}
    \label{tab:aba_motion}
\end{table}
\begin{table}[t]\scriptsize
\centering
	\renewcommand{\arraystretch}{1.1}
	\caption{\textbf{Ablation studies of image synthesis.}}
	\begin{minipage}{0.45\textwidth}%
	\centering
    \setlength\tabcolsep{0.6mm}
    \begin{tabular}{cc|cccc}
        \toprule

          Inp. BG & FG Comp.

 &  ${\rm FID}$& ${\rm AKD}$& ${\rm AED(C)}$& ${\rm HPD(C)}$ \\
        \midrule
        ~ & ~ & 30.1 & 1.54 & 7.3  &  0.14\\ 
        \rowcolor{gray!20}
         \checkmark & ~ & 29.2 & 1.30 & 2.74  &  0.13\\

           ~& \checkmark  & 30.0 & 1.52 & 5.70  &  0.13\\ 
           \checkmark & \checkmark & 29.7 &1.47& 10.0 & 0.13  \\ 
         
        \bottomrule
    \end{tabular}
    \end{minipage}
    \hspace{5.65em}
    \begin{minipage}{0.4\textwidth}%
    \setlength\tabcolsep{0.4mm}
    \begin{tabular}{l|cccc}
        \toprule
        \#Views & ${\rm FID}$& ${\rm AKD}$& ${\rm AED(C)}$& ${\rm HPD(C)}$ \\
	    \midrule
        No Ref. & 34.2 & 2.53 & 3.21  &  0.13\\

        2 & 31.3 & 1.53 & 3.07  &  0.13\\ 
        \rowcolor{gray!20}
        4 & 29.2 & 1.30 & 2.7  &  0.13\\ 
        8 & 30.2 & 1.31 & 2.1  &  0.13\\ 
        \bottomrule
    \end{tabular}
    \end{minipage}
    \label{tab:aba_app}
\end{table}
\begin{table}[t]\scriptsize
	\renewcommand{\arraystretch}{1.1}
	\caption{\textbf{Ablation studies of training datasets.}
	}
	\centering
    \setlength\tabcolsep{0.7mm}
    \begin{tabular}{c|ccccccc}
        \toprule
        Method & ${\rm FID\downarrow}$&${\rm PSNR\uparrow}$&${\rm SSIM\uparrow}$& ${\rm AKD\downarrow}$& ${\rm APD(C)\downarrow}$& ${\rm AED(C)\downarrow}$&${\rm CSIM(C)\uparrow}$ \\
	    \midrule
        \rowcolor{gray!20}
         Full Datasets & 29.2 & 26.1 & 85.9& 1.30 & 2.7 & 0.13 & 39.2 \\ 
         remove VoxCelebvHQ & 32.5 &26.0 & 85.8& 1.43 & 2.9& 0.13&37.7  \\ 
         remove VFHQ & 37.1 & 25.4& 84.3 &1.49 & 3.6& 0.13&38.5  \\ 
	    \bottomrule
    \end{tabular}
    \label{tab:dataset}
\end{table}

\subsection{Comparisons with SOTA methods}
In this section, we contrast MobilePortrait's video-driven performance with other techniques in Table~\ref{tab:main_exp}, and will later include an audio-driven comparison. Given that audio-driven methods often use video-driven approaches for rendering, video-driven analysis serves as a reliable measure of synthesis quality. In same-id scenarios, the source image is sampled from the driving video, meaning there exists ground truth video for reference. In cross-id scenarios, the source image is not derived from the driving video; instead, we randomly select and sample a frame from another video in datasets, so there is no GT video for direct comparison, and we do not assess reference-based image quality metrics. 

It can be discerned that MobilePortrait, despite employing a smaller computational load, achieves outcomes comparable to those with greater computational resources and excels in key metrics, leading in AKD and BCI and ranking second in FID, which assess motion and image quality. Furthermore, during cross-identity reenactment, the lead in HPD, AED and BCI metrics also demonstrates the effectiveness of MobilePortrait. While MobilePortrait does not achieve the best results in the CSIM, later visualization results show that it can yield satisfactory outcomes.

\subsection{Comparisons among Different Computational Loads}
Here, a comparative analysis of performance across various computational scales (FLOPs) is provided in right part of Table~\ref{tab:gflops_exp}. By reducing the number of channels and layers, we obtain models of different sizes. MobilePortrait remarkably maintains satisfactory performance on key metrics such as FID and AKD, as well as cross-identity motion accuracy like HPD and AED, even when computational resources are limited to just 4 GFLOPs, marking a significant improvement over the baseline, which does not incorporate external facial and appearance knowledge as demonstrated in Figure~\ref{fig:framework}. Moreover, visualization results in Figure~\ref{fig:gflops_vis} showcasing our approach's effectiveness. Concurrently, in the left part of Table~\ref{tab:gflops_exp} also details the computational resource consumption of MobilePortrait on mobile devices, underscoring its efficient viability on mobile platforms. 
\subsection{Ablation Studies}
\textbf{Motion Generation.} we conduct ablation experiments to validate the effectiveness of proposed components. We assess key metrics for image quality and motion, such as FID and AKD, along with AED and HPD specifically in cross-identity scenarios, denoted as AED(C) and HPD(C). Table~\ref{tab:aba_motion} presents comparisons among employing mixed keypoints, neural keypoints only and face keypoints only settings, where mixed keypoints demonstrate significant performance improvements. Additionally, excluding our proposed facial knowledge losses degrades results. We also explored alternative approaches to integrating NK and FK beyond the mixed keypoint predictor. For instance, as shown in (b) of Figure~\ref{fig:demo_motion} and drawing inspiration from literatures~\cite{siarohin2019fomm,wang2021facev2v,siarohin2021mraa}, we perform fusion on the initial transformation. We transform FK into sparse motions to generate transformations, which, when concatenated with NK's transformations, yield a combined transformations. Alternatively, in (c) of the figure, both NK and FK are converted into heatmaps, which are then directly fed into a convolutional network to generate optical flow. However, these methods did not achieve better motion accuracy than mixed keypoints. Additionally, we experimented with removing the residual optical flow and observed that it indeed resulted in decreased motion accuracy, although it also introduced some perturbations to the FID. The experimental results demonstrate that integrating explicit and implicit information significantly improves the generated outcomes in terms of image quality and motion, while the fusion form of mixed keypoints is a simple and effective design.

\begin{table}[t]\scriptsize
	\centering
	\caption{\textbf{Comparisons with audio-driven methods} }
    \setlength\tabcolsep{0.65mm}{
    \begin{tabular}{l|cccc}
        \toprule
        
        Method& Sync-C$\uparrow$ & Sync-D$\downarrow$  &BSI$\uparrow$& Training Data \\
        \midrule
        MakeItTalk~\cite{zhou2020makelttalk}  & 4.77 &10.19&98.0& 109 ID \\
    SadTalker~\cite{zhang2023sadtalker} &7.32&7.87& 98.2& 1890 Videos, 46 ID\\
    Real3D~\cite{ye2024real3d} & 7.06&7.77&98.0 &200 Hours, 6000 ID \\
    \rowcolor{gray!20}
    MobilePortrait & 6.01 & 9.02 &\textbf{98.5}& 16 Hours, 20 ID \\
        
        \bottomrule
    \end{tabular}}
    \label{tab:audio}
\end{table}

\textbf{Enhanced Background Synthesis.} In this section, we assess the effectiveness and usage of pseudo background in synthesis networks and list experimental results the left part of Table~\ref{tab:aba_app}, where the setting employed in our method is marked in gray. We investigated four configurations, namely whether pseudo background should be input into the synthesis network (abbreviated as Inp. BG) and whether the model should synthesize the background (given the presence of pseudo backgrounds, we have the option to generate only the foreground and then composite it onto the background by additionally predicting an alpha channel, abbreviated as FG Comp.). We find that pseudo-background integration indeed enhances performance by transforming the task from full generation to knowledge-aided synthesis. Table~\ref{tab:aba_app} shows that separately generating and merging the foreground and background does not significantly improve performance. Our method, highlighted in row two, enhances synthesis by end-to-end training with pseudo backgrounds derived from driving images, enabling effective utilization of this knowledge in creating the final image. Real3D, which employs volume rendering, attempts to integrate the rendered head with the background using a split-and-merge-like strategy, but this can lead to inconsistent motion and visible discrepancies, as depicted in Figure~\ref{fig:firstpage} and Figure~\ref{fig:demo_vis}.

\par
\textbf{Enhanced Foreground Synthesis.} For the pseudo multi-view inputs, experiments are conducted to examine the influence of different numbers of pseudo multi-view inputs on synthesis quality. We examine configurations with 0 (indicating the absence of multiview inputs), 2, 4, and 8 multiview inputs, and observe that each incremental addition of multiview inputs proportionately enhances the synthesis outcomes. In the right part of Table~\ref{tab:aba_app}, as the number of images increases, the improvement in results begins to saturate, an observation consistent with phenomena encountered in video classification~\cite{tsn,feichtenhofer2019slowfast} and 3D-related tasks~\cite{su2015multi}. This correlation is striking. Indeed, as multiview inputs near the frame count of the driving video, the synthesis process naturally evolves from generating static images to creating dynamic video sequences. Adding temporal data improves stability, yielding predictable and notable enhancements.
\par
The experiments confirm our method's effectiveness and our premise that external knowledge boosts the motion and synthesis networks with minimal extra computational cost, allowing model to deliver satisfactory results with reduced overhead.

\textbf{Training data.} In addition to the method, we are also interested in whether the data plays a significant role in achieving satisfactory performance for lightweight neural head avatar methods. We sequentially remove training datasets, starting with the relatively lower-quality Dataset VoxCeleb2~\cite{vox}, followed by the removal of the high-quality Dataset VFHQ~\cite{xie2022vfhq}, leaving only Dataset CelebvHQ~\cite{zhu2022celebv}. Table~\ref{tab:dataset} illustrates that the removal of datasets leads to a decline in performance. However, this decline is mainly reflected in the image quality index FID, with a relatively smaller impact on motion accuracy. This results demonstrate the robust motion modeling capability of our motion generation module, which, even with less data, maintains superior motion accuracy compared to some methods listed in Table~\ref{tab:main_exp}. The VFHQ dataset, due to its higher clarity, has a more pronounced impact on FID, aligning with the expectations.

\subsection{Experimental Results on More Application Scenarios
}
\par
\textbf{Comparisons among Audio-to-Motion.} We compared MobilePortrait with some audio-driven methods. We used the 100 videos from HDTF for testing and measured lip synchronization using the Sync-D and Sync-C metrics generated by SyncNet~\cite{prajwal2020wav2lip} and evaluated background consistency using BSI. The results in Table~\ref{tab:audio} indicate that MobilePortrait achieves comparable performance to some audio-driven methods, outperforms one of them, and exhibits superior visual stability. It is noteworthy that MobilePortrait's primary focus is to achieve a real-time neural head avatar method on mobile devices. Our audio-to-motion module, trained on a limited set of our own speaking videos, serves as a baseline to showcase its adaptability to audio inputs. We provide audio-driven video samples in the supplementary materials, demonstrating that MobilePortrait can achieve satisfactory results.

\par
\begin{figure}[!ht]
    \centering
    \includegraphics[width=1.0\linewidth]{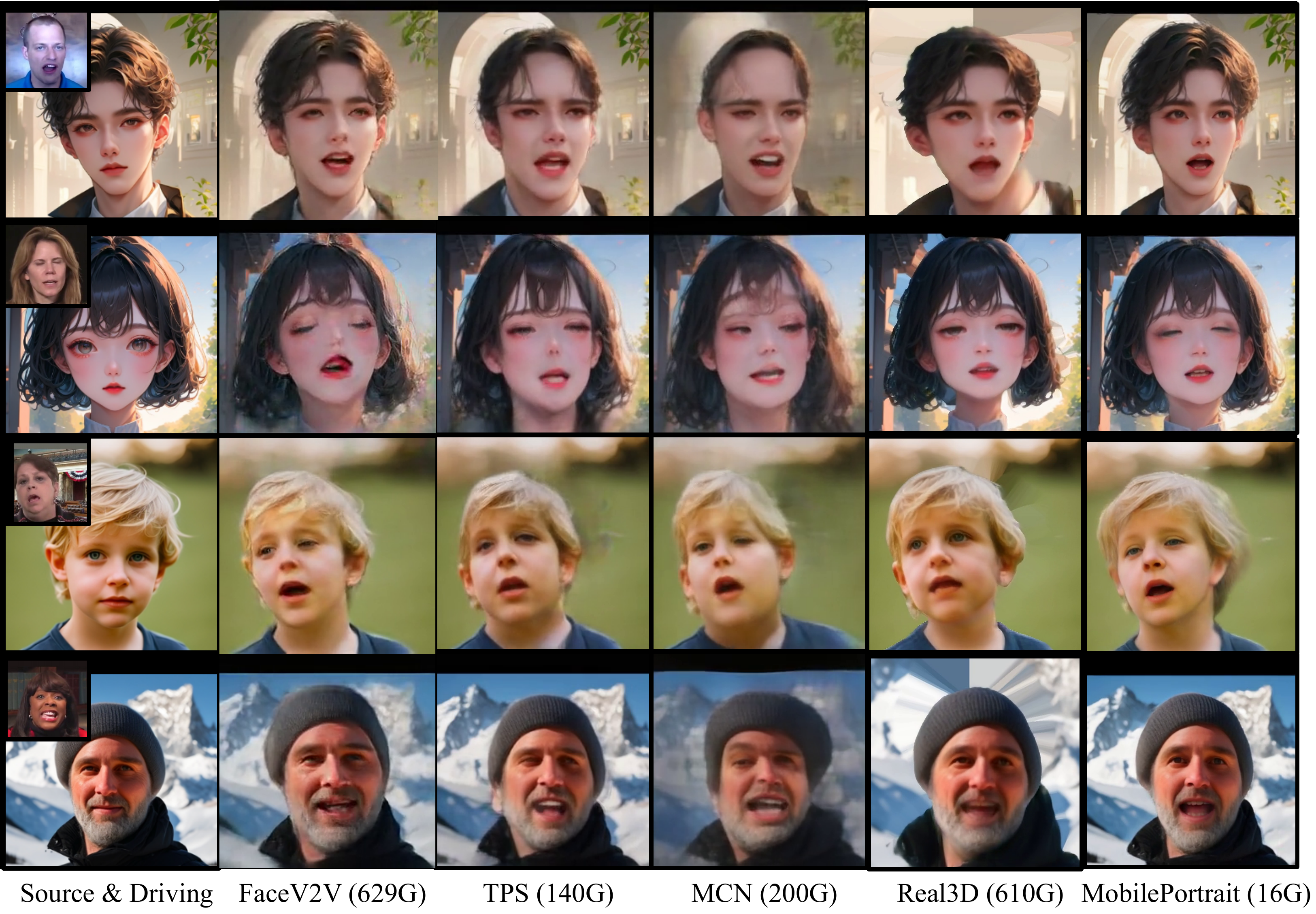}

    \caption{
    \textbf{Visualization Results.} To compare with other methods visually, we selected various styles of input images and rich motions to demonstrate the robustness of MobilePortrait (more results are shown on the project page). The video results are provided in the supplementary materials.
    }
    \label{fig:demo_vis}
\end{figure}
\textbf{Robustness to Motion and Appearance.} To further demonstrate the utility of MobilePortrait, in Figure~\ref{fig:demo_vis}, we provide a visual analysis validating its robustness in cross-identity reenactment,  with challenging scenarios like non-real images, complex backgrounds and large motions. Current state-of-the-art methods, when confronted with these challenging cases, reveal numerous unfavorable results. With the help of external knowledge, MobilePortrait achieves satisfactory results with significantly less computational effort.

\section{Conclusion}
In this work, we address the overlooked challenge of creating lightweight one-shot neural head avatars and introduce MobilePortrait, to the best of our knowledge, the first real-time solution for mobile devices. By employing a mixed representation of explicit and implicit keypoints, along with pseudo multiview and background, we enhance the network's motion generation and synthesis capabilities with external knowledge, enabling MobilePortrait to achieve neural head avatars with simple lightweight U-Nets. Extensive experiments confirm that MobilePortrait matches state-of-the-art performance in synthesis quality and motion accuracy, and supports both video and audio driving inputs.

\clearpage 

\bibliographystyle{splncs04}
\bibliography{main}
\end{document}